\documentclass[10pt,twocolumn,letterpaper]{article}

\usepackage{iccv}
\usepackage{times}
\usepackage{epsfig}
\usepackage{graphicx}
\usepackage{amsmath}
\usepackage{amssymb}
\usepackage[accsupp]{axessibility}

\usepackage{booktabs}
\usepackage{caption}
\usepackage{subcaption}
\usepackage[export]{adjustbox}
\usepackage{hhline}
\usepackage{multirow}
\usepackage[table,xcdraw]{xcolor}
\usepackage{makecell}
\usepackage[inline, shortlabels]{enumitem}
\usepackage{bbm}
\usepackage{bm}
\usepackage{cite}
\usepackage{kotex}
\usepackage{comment}
\include{mymacros}

\usepackage[breaklinks=true,bookmarks=false,colorlinks]{hyperref}

\usepackage[capitalize]{cleveref}
\crefname{section}{Sec.}{Secs.}
\Crefname{section}{Section}{Sections}
\Crefname{table}{Table}{Tables}
\crefname{table}{Tab.}{Tabs.}

\iccvfinalcopy 

\def\iccvPaperID{4444} 

\ificcvfinal\fi

\begin{document}

\title{EP2P-Loc: End-to-End 3D Point to 2D Pixel Localization\\for Large-Scale Visual Localization}

\author{Minjung Kim \quad Junseo Koo \quad Gunhee Kim\\
Seoul National University\\
{\tt\small \{minjung.kim, junseo.koo\}@vision.snu.ac.kr, gunhee@snu.ac.kr}\\
{\tt\small \href{https://github.com/minnjung/EP2P-Loc}{https://github.com/minnjung/EP2P-Loc}}
}

\maketitle
\ificcvfinal\thispagestyle{empty}\fi

\begin{abstract}

Visual localization is the task of estimating a 6-DoF camera pose of a query image within a provided 3D reference map.
Thanks to recent advances in various 3D sensors, 3D point clouds are becoming a more accurate and affordable option for building the reference map, 
but research to match the points of 3D point clouds with pixels in 2D images for visual localization remains challenging.
Existing approaches that jointly learn 2D-3D feature matching suffer from low inliers due to representational differences between the two modalities, and the methods that bypass this problem into classification have an issue of poor refinement.
In this work, we propose EP2P-Loc, a novel large-scale visual localization method that mitigates such appearance discrepancy and enables end-to-end training for pose estimation.
To increase the number of inliers, we propose a simple algorithm to remove invisible 3D points in the image, and find all 2D-3D correspondences without keypoint detection.
To reduce memory usage and search complexity, we take a coarse-to-fine approach where we extract patch-level features from 2D images, then perform 2D patch classification on each 3D point, and obtain the exact corresponding 2D pixel coordinates through positional encoding.
Finally, for the first time in this task, we employ a differentiable PnP for end-to-end training. 
In the experiments on newly curated large-scale indoor and outdoor benchmarks based on 2D-3D-S and KITTI, we show that our method achieves the state-of-the-art performance compared to existing visual localization and image-to-point cloud registration methods.

\end{abstract}

\section{Introduction}

\begin{figure}[t]
  \centering
   \includegraphics[width=1.0\linewidth]{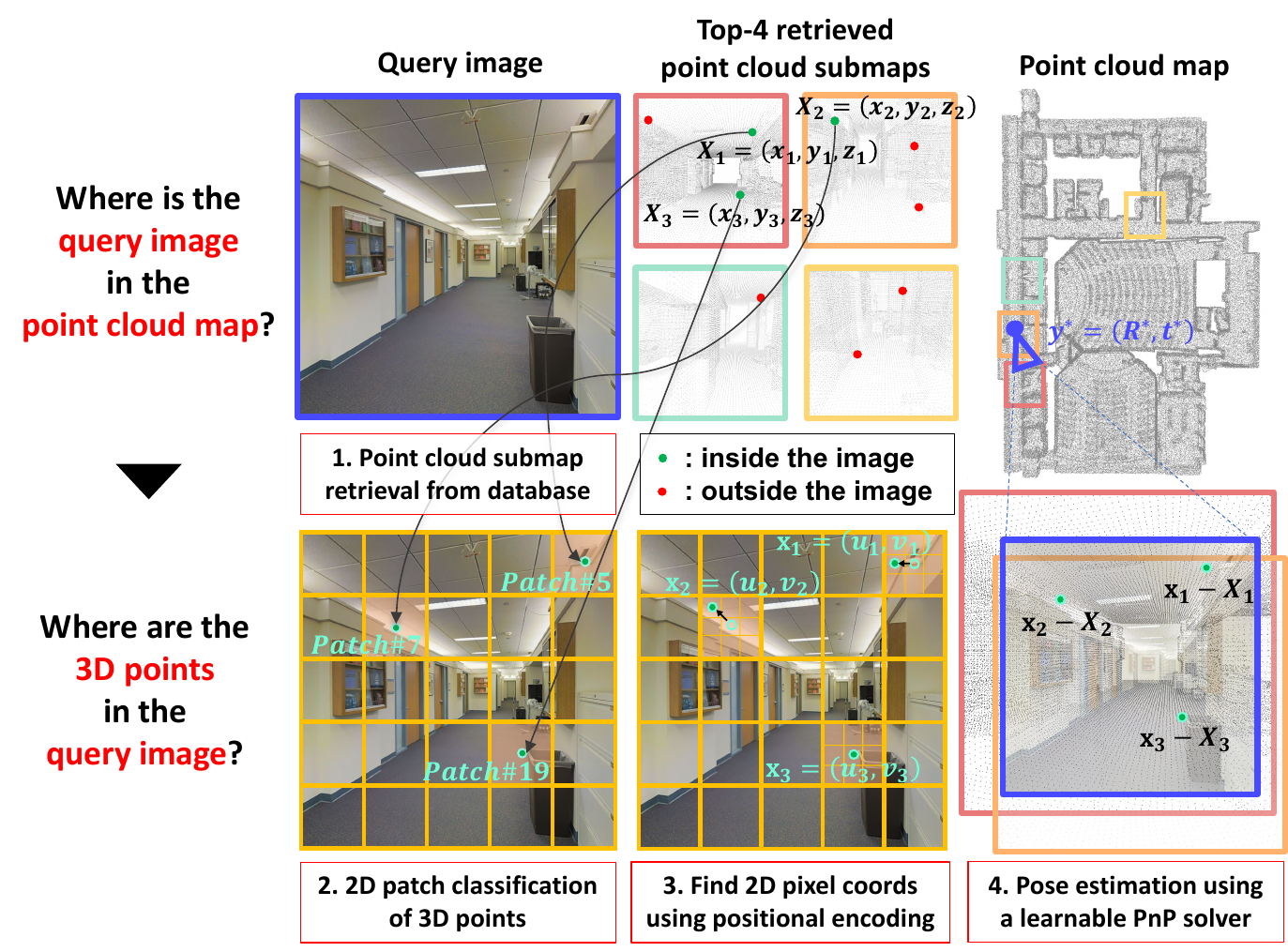}
   \caption{Visual localization of EP2P-Loc. For a query image, we (1) retrieve top-$4$ point cloud submaps from the database, (2) perform 2D patch classification for each 3D point in the retrieved submaps, (3) find precise 2D pixel coordinates using a positional encoding, and (4) estimate the camera pose of the image using a learnable PnP solver \cite{epropnp}.}
   \label{fig:pipeline}
\end{figure}

Visual localization \cite{netvlad, featloc, d2net, kapture, PCLoc, inloc, hfnet} aims at estimating the precise $6$-degree-of-freedom (DoF) camera pose (\ie position and orientation) of a given 2D image within a 3D reference map. 
Since it can position the pose of a photo-taker in an environment without localization information such as GPS,
it has a variety of applications including autonomous driving, robot navigation, and augmented reality. 
Previous works use the Perspective-n-Point (PnP) algorithm \cite{pnpsolver} to compute the camera pose of a query image based on the coordinates in the 3D reference map.
The 3D reference map is typically built from a collection of images using Structure-from-Motion (SfM) reconstruction \cite{sfm, sfm1, sfm2, sfm4, colmap}, 
and the associated keypoints and descriptors are stored with the map in the form of point clouds to establish 2D-3D correspondences. 
Therefore, the accuracy of visual localization is highly dependent on the quality of the 3D reference map,
and there are multiple challenges, including varying lighting, weather and seasonal conditions in outdoor environments, and repetitive patterns and insufficient textures in indoor environments. 

With the recent advance in 3D sensors, such as RGB-D scanners and LiDARs, it is possible to generate large-scale 3D point cloud maps with no help from SfM. 
This could be a more practical approach compared to SfM since the point clouds are denser and more accurate. 
However, visual localization has difficulty in matching descriptors from 2D images with those from 3D reference point clouds due to the big discrepancy between 2D and 3D representations. 
This is because 2D images are represented in the form of a grid of colors and texture, while 3D point clouds are relatively sparse and lack texture information due to empty spaces in the 3D space.

To tackle this problem, several recent studies have tried to match 2D pixels of an image with 3D points of a point cloud.
2D3D-MatchNet \cite{2d3dmatchnet} is a pioneer work 
that proposes a deep neural network to embed the descriptors of 2D and 3D keypoints into a shared feature space.
However, it still suffers from a low inlier problem, which means that the number of predicted 2D-3D correspondences used for pose estimation is not enough, since it is difficult to mitigate the severe appearance differences between the two modalities.
P2-Net \cite{p2net} detects keypoints and per-point descriptors based on the RGB-D scan datasets \cite{7Scenes1, 7Scenes2}, but every 2D pixel must have a corresponding 3D point.
Moreover, obtaining features from all 2D pixels and 3D points can be inefficient in terms of memory usage and the amount of computation for the visual localization in large-scale environments.
DeepI2P \cite{deepi2p} replaces descriptor matching with classification to reduce computation while increasing accuracy.
However, it is limited for large-scale visual localization as will be shown in our experiments later, since it is challenging to find exact 2D-3D correspondences from the grid classification of each 3D point.

In this work, we propose \textit{EP2P-Loc}, a novel approach to large-scale visual localization that prevents the low inlier problem by processing all 2D pixels in the query image and 3D points in the 3D reference maps, while reducing memory usage and search space.
\Cref{fig:pipeline} outlines the pipeline of our method at inference.
Initially, the 3D point cloud reference map is divided into a set of submaps for effective candidate selection.
Like a coarse-to-fine approach, we retrieve relevant 3D point cloud submaps from the database for a query image.
We then perform 2D patch classification to determine to which patch of the image the 3D points in the retrieved submaps belong. 
Next, we find the exact 2D pixel coordinates of the 3D points using positional encoding, instead of storing features for all 2D pixels. 
Finally, the obtained 2D-3D correspondences pass into the PnP layer to estimate the camera pose of the image.

The key novelty of our approach lies in its ability to effectively learn features of both 2D pixels and 3D points while not only handling invisible 3D points in 2D images using our \textit{Invisible 3D Point Removal (IPR)} algorithm but also finding all 2D-3D correspondences without the need for keypoint detection in a coarse-to-fine manner, resulting in a higher number of inliers.
Moreover, for the first time in this task, we adopt an end-to-end learnable PnP solver \cite{epropnp} for high-quality $6$-DoF pose estimation.
We demonstrate that this approach is more accurate and efficient than selecting top $k$ pairs and manually putting them into the PnP solver.
For empirical validation in large-scale indoor and outdoor environments, we establish benchmarks based on the open-source Stanford 2D-3D-Semantic (2D-3D-S) \cite{2D-3D-S} and KITTI \cite{kitti} datasets,
and show the state-of-the-art performance compared to existing image-based visual localization methods \cite{netvlad, dsac_star, superpoint, d2net, kapture, PCLoc, hfnet, superglue} and image-to-point cloud methods \cite{2d3dmatchnet, deepi2p}.
Our method uses only the coordinate values of the 3D point cloud and assumes that every 3D point is not associated with a pixel in the 2D image and vice versa. 
Through experiments, we also demonstrate that our method is applicable to various 3D global point cloud maps generated in different ways, including 3D LiDAR data, RGB-D scans, and SfMs.

\section{Related work}

\begin{figure*}[t]
  \centering
   \includegraphics[width=1.0\linewidth]{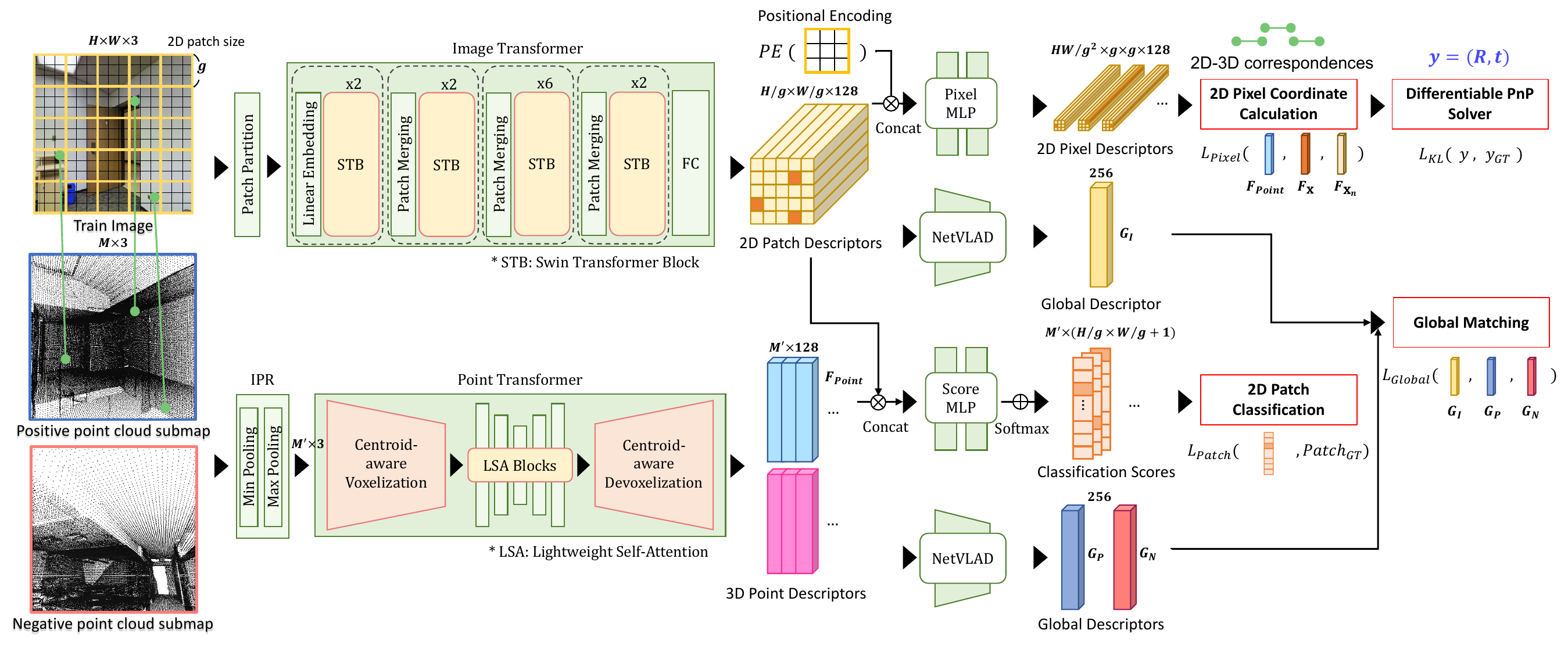}
   \caption{Overall architecture of our EP2P-Loc network.}
   \label{fig:model_details}
\end{figure*}

\textbf{Visual localization.}
Visual localization is the problem of estimating the $6$-DoF camera pose of a query image within a provided 3D reference map.
Image retrieval-based approaches \cite{netvlad, kapture, delf, hfnet} estimate the location by retrieving the most similar image from a geo-tagged database.
The retrieved location can limit the scope of the search area within the large 3D reference map generated by SfM \cite{sfm, sfm1, sfm2, sfm4, colmap}, 
and thus these methods allow to speed up the search in a large environment.
Structure-based methods  \cite{sfm, kapture, sfm1, sfm2, hfnet, sfm4, sattler2016efficient, sfm3, inloc} extract 2D local descriptors \cite{superpoint, d2net, sift, r2d2} from database images, construct the 3D reference map through SfM, and store one or more local descriptors at each 3D point.
Given the query image, they extract 2D local descriptors, match local descriptors stored in the 3D reference map to find 2D-3D correspondences, and predict the $6$-DoF camera pose from top $k$ pairs using the PnP solver \cite{epnp} with RANSAC \cite{pnpsolver}.
Some structure-based methods \cite{hfnet, inloc} adopt image retrieval-based methods to reduce search space for speed-up.
However, these methods heavily rely on image-based features, and thus are not robust to various lighting, weather and seasonal conditions in outdoor environments, and repetitive patterns and insufficient textures in indoor environments.
Some methods \cite{featloc, mapnet, posenet_variant0, hourglass-pose, apr} directly regress the camera pose from the single query image, but are not competitive in terms of accuracy.
PCLoc \cite{PCLoc} and FeatLoc \cite{featloc} reduce viewpoint differences between the query and the database, and estimate the pose by synthesizing new views with RGB-D images.
Although they can reduce translational errors, their synthesizing modules require dense point clouds or RGB-D images, which limit the training setup or dataset selection.

\textbf{Image-to-point cloud registration.}
To estimate the relative pose between an image and a point cloud,
2D3D-MatchNet \cite{2d3dmatchnet} and LCD \cite{lcd} propose deep networks to jointly learn the descriptors from the 2D image patches and 3D point cloud patches.
However, they are limited in that keypoints must be set in advance, and in particular, 2D3D-MatchNet, which is the only method that aims at the same task as ours, suffers from a low inlier problem due to the difference in representation between the two modalities.
3DTNet \cite{3dtnet} learns 3D local descriptors that take 2D and 3D local patches as inputs while using the 2D features as auxiliary information only.
Cattaneo \etal \cite{2d3dmatch1} creates a shared embedding space between 2D images and 3D point clouds adopting a teacher/student model.
However, they still require 2D and 3D pre-defined keypoints, leading to inaccurate matches without consistent keypoint selection.
P2-Net \cite{p2net} jointly learns local descriptors and detectors for pixel and point matching, but this is applicable only when every 2D pixel has a corresponding 3D point based on RGB-D scans. 
DeepI2P \cite{deepi2p} transforms the 2D-3D feature matching problem into a classification problem of whether a projected point lies within an image frustum.
However, this method is limited for large-scale visual localization tasks because it cannot find exact 2D-3D correspondences and uses both an image and a point cloud as inputs to estimate the relative pose between them.

\section{EP2P-Loc}

Our goal is to predict the $6$-DoF camera pose of a query 2D image within a 3D global reference map of a point cloud.
We assume that the global map is partitioned into a set of submaps, which will be described in \Cref{sec:datasets}.
During training, we are given the image $I \in \mathbb{R}^{H \times W \times 3}$, 
the positive point cloud submap $P=\left\{\left(x_{i}, y_{i}, z_{i}\right)\right\}_{i=1}^M \in \mathbb{R}^{M \times 3}$ where $M$ is the number of 3D points, 
the set of negative point cloud submaps $N=\left\{N_{i}\right\}_{i=1}^{\left\lvert N \right\rvert}$ where $N_{i} \in \mathbb{R}^{M \times 3}$,
the camera pose $y_{gt} = (\mathbf{R}, \mathbf{t}) \in SE(3)$, 
and the camera intrinsic matrix $\mathbf{K} \in \mathbb{R}^{3 \times 3}$.
The positive submap indicates the one that contains the point clouds of the scene in the image, while the negative submaps are randomly taken from other regions. 
\Cref{fig:pipeline} describes the overall pipeline of our approach during inference, 
and \Cref{fig:model_details} illustrates our network architecture for end-to-end training.
We below explain its details in the training and inference process. 
The implementation details are presented in \Cref{sec:implementation_details}.

\subsection{Feature extraction} \label{sec:feature_extraction}
Since we follow the coarse-to-fine matching scheme, we extract both local and global descriptors from both image and point cloud submaps.
Our feature extraction network for images and point clouds are 
respectively based on the Swin Transformer \cite{swin} and the Fast Point Transformer \cite{fast_point_transformer}.
We use Transformer-based extractors~\cite{attention}, 
since they are known to be robust to modality differences, making it possible to extract features that share the embedding space regardless of input modality \cite{clip, unicl}. 
Then, the global descriptors are obtained by aggregation of local descriptors.
These global descriptors are used to reduce search space from the 3D global reference map by retrieving relevant submaps from the database, which we call \textit{global matching}.


\textbf{2D image descriptors}. 
We first divide the input image $I$ into tokens of the size of $4 \times 4$. 
For feature extraction of each 2D patch, we use the modified Swin-S Transformer with a window size of $32$, which consists of $4$-stage Swin-T Transformer blocks and an FC layer. 
The output resolution is then $\frac{H}{g} \times \frac{W}{g}$, where $g$ is our 2D patch size,
and we get a descriptor of $128$-dim for each 2D patch.
The global descriptor is obtained by applying a NetVLAD \cite{netvlad} layer to the FC layer. 
Finally, we obtain a $256$-dim global descriptor $G_{I}$ and $128$-dim 2D patch descriptors.

\begin{figure}[t]
  \centering
  \begin{subfigure}{.15\textwidth}
    \centering
    \includegraphics[width=.98\textwidth]{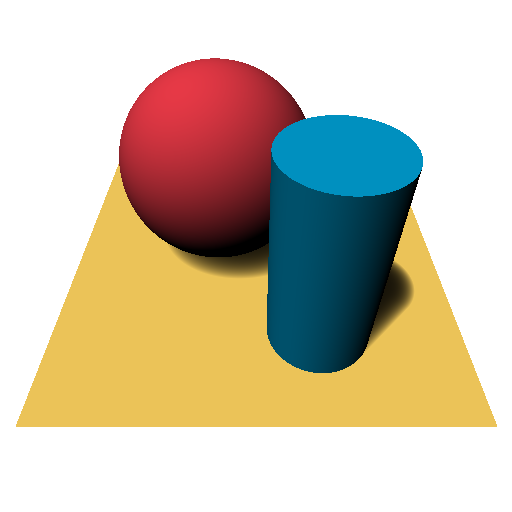}
    \includegraphics[width=.98\linewidth]{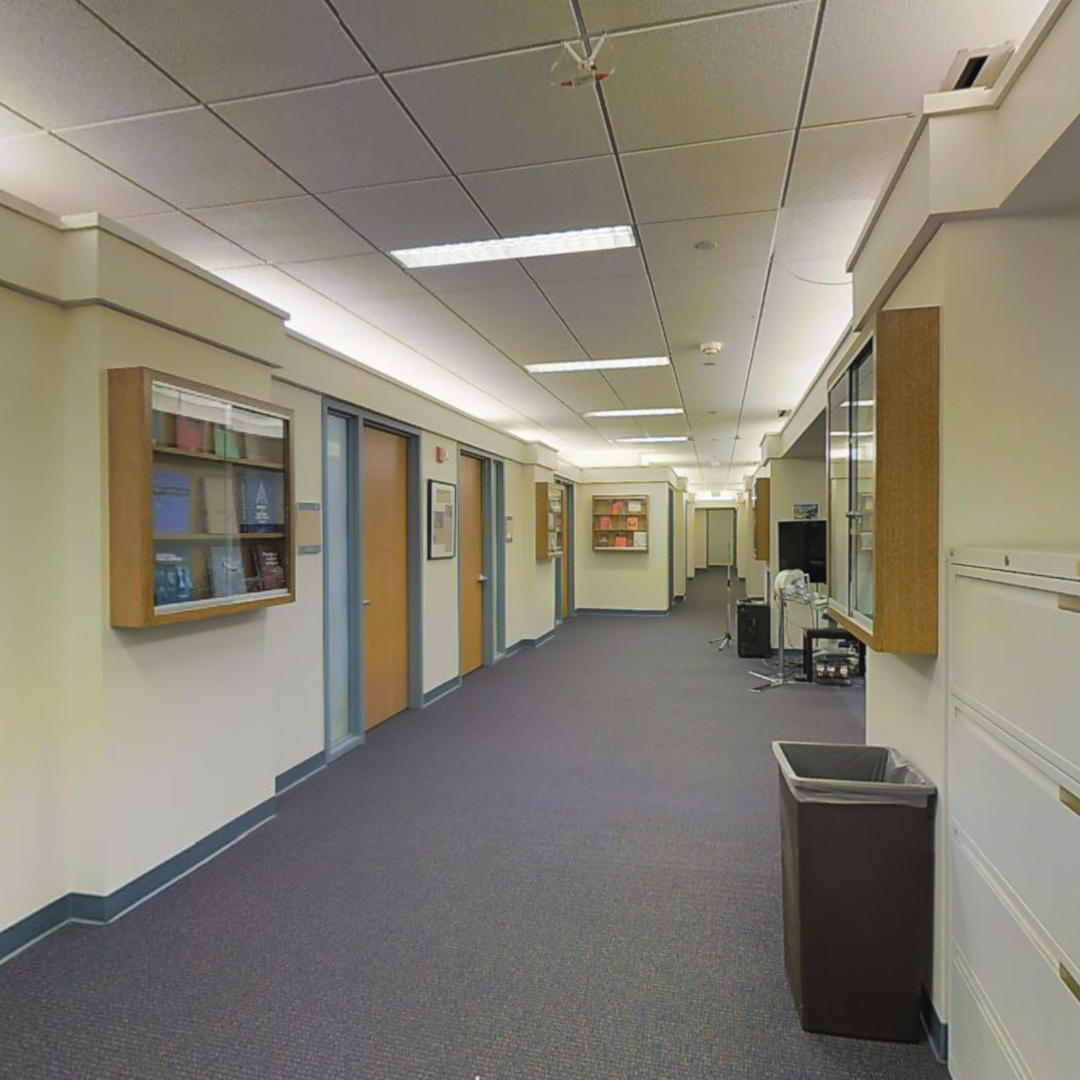}
    \includegraphics[width=.98\linewidth]{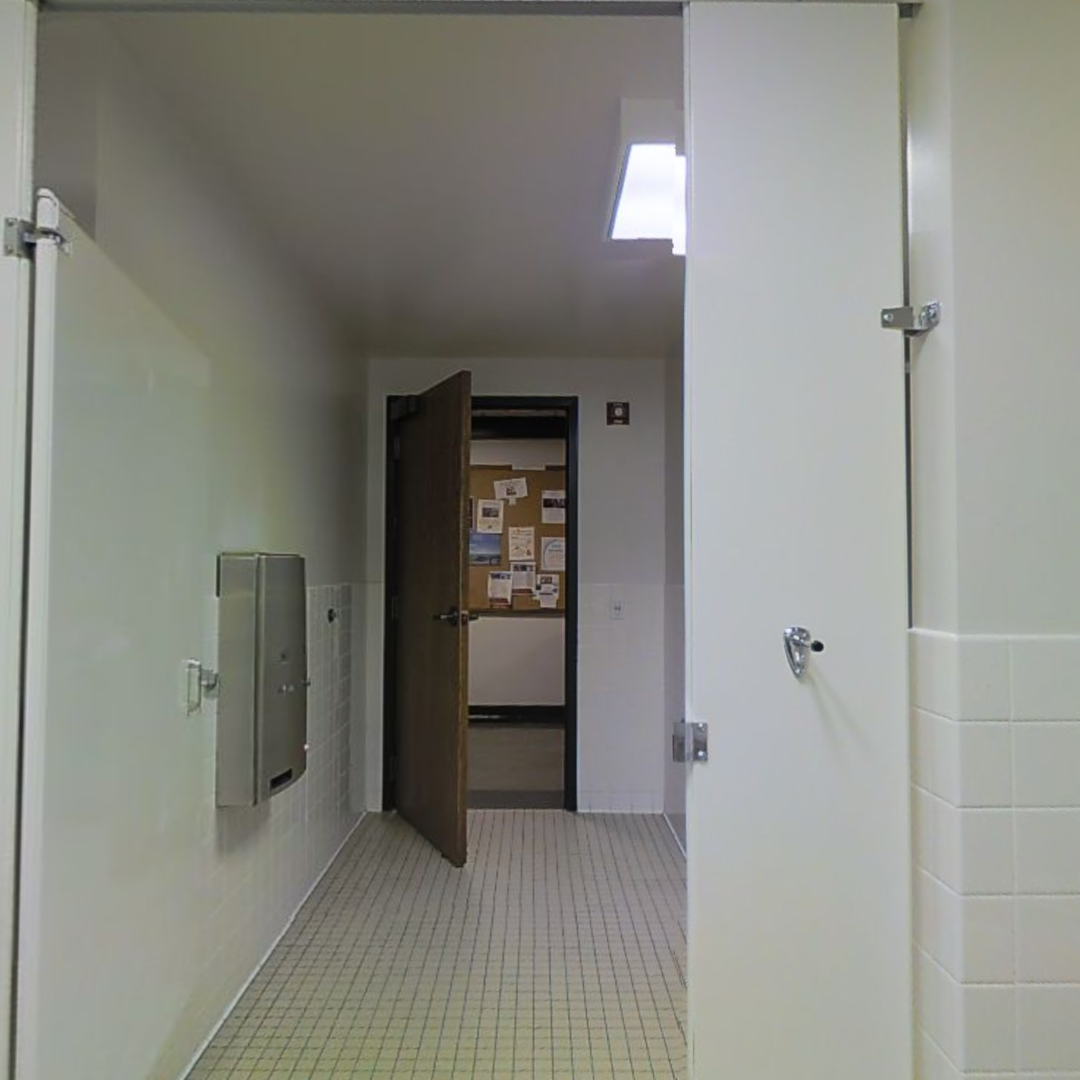}
    \caption{Images}
    \label{fig:2d3ds_rgb}
  \end{subfigure}
  \begin{subfigure}{.15\textwidth}
    \centering
    \includegraphics[width=.98\textwidth]{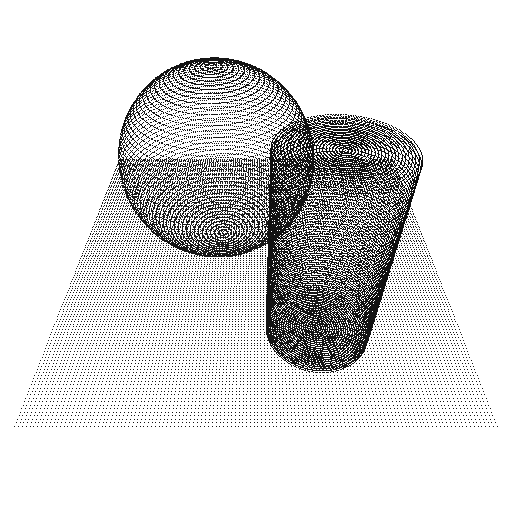}
    \includegraphics[width=.98\linewidth]{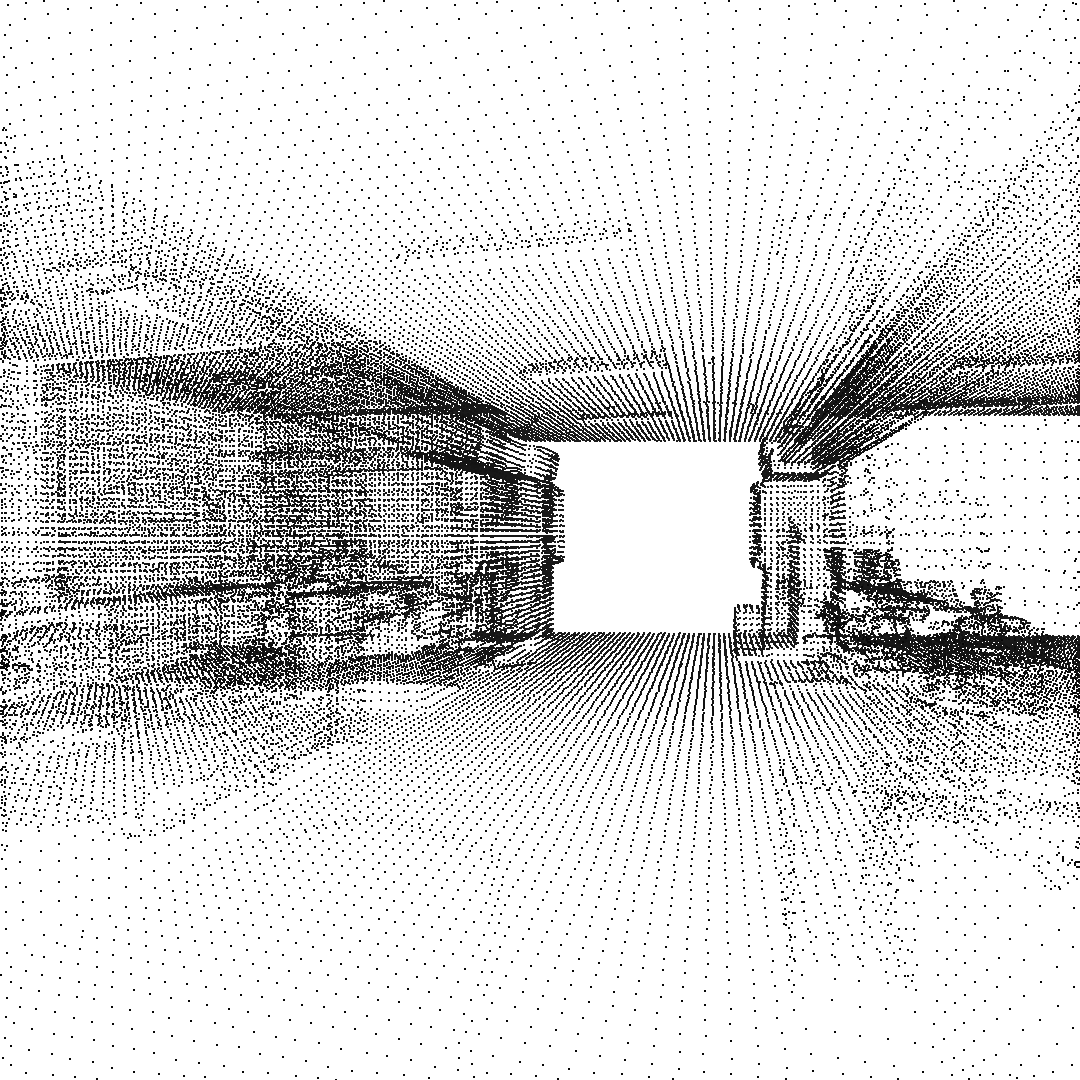}
    \includegraphics[width=.98\linewidth]{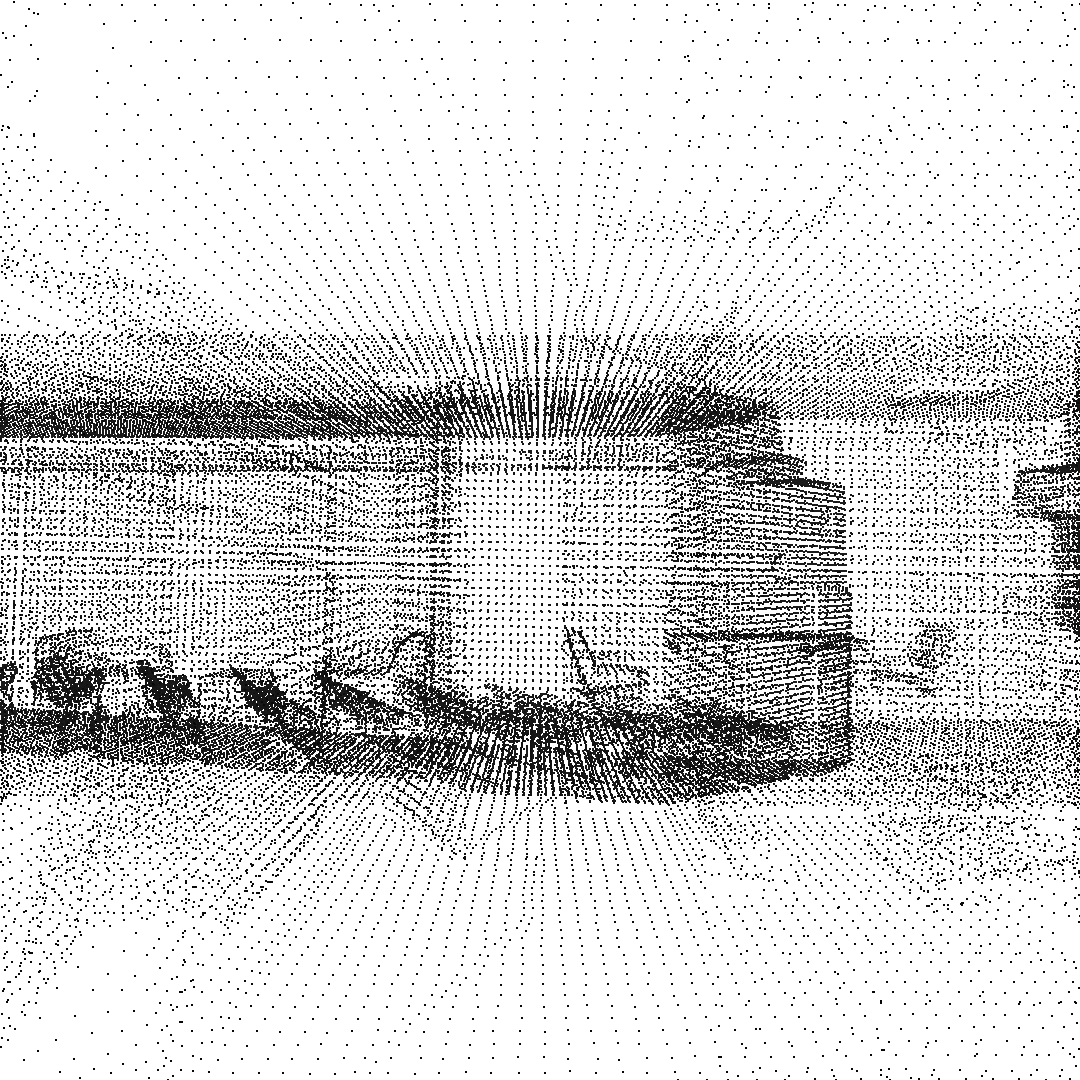}
    \caption{Before removal}
    \label{fig:2d3ds_clean_before}
  \end{subfigure}
  \begin{subfigure}{.15\textwidth}
    \centering
    \includegraphics[width=.98\textwidth]{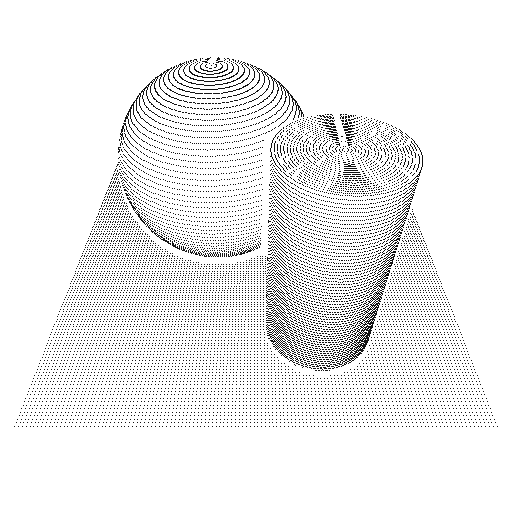}
    \includegraphics[width=.98\linewidth]{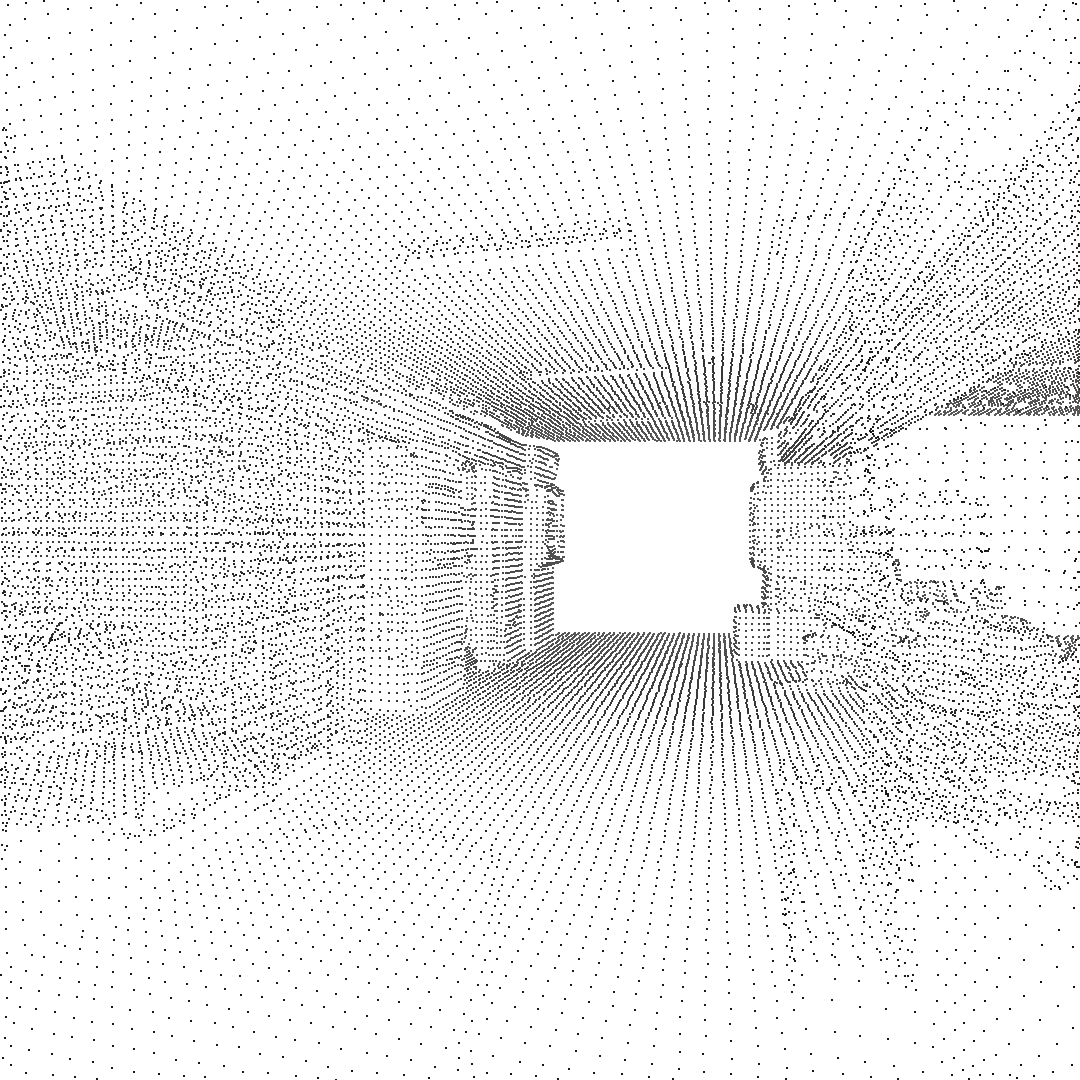}
    \includegraphics[width=.98\linewidth]{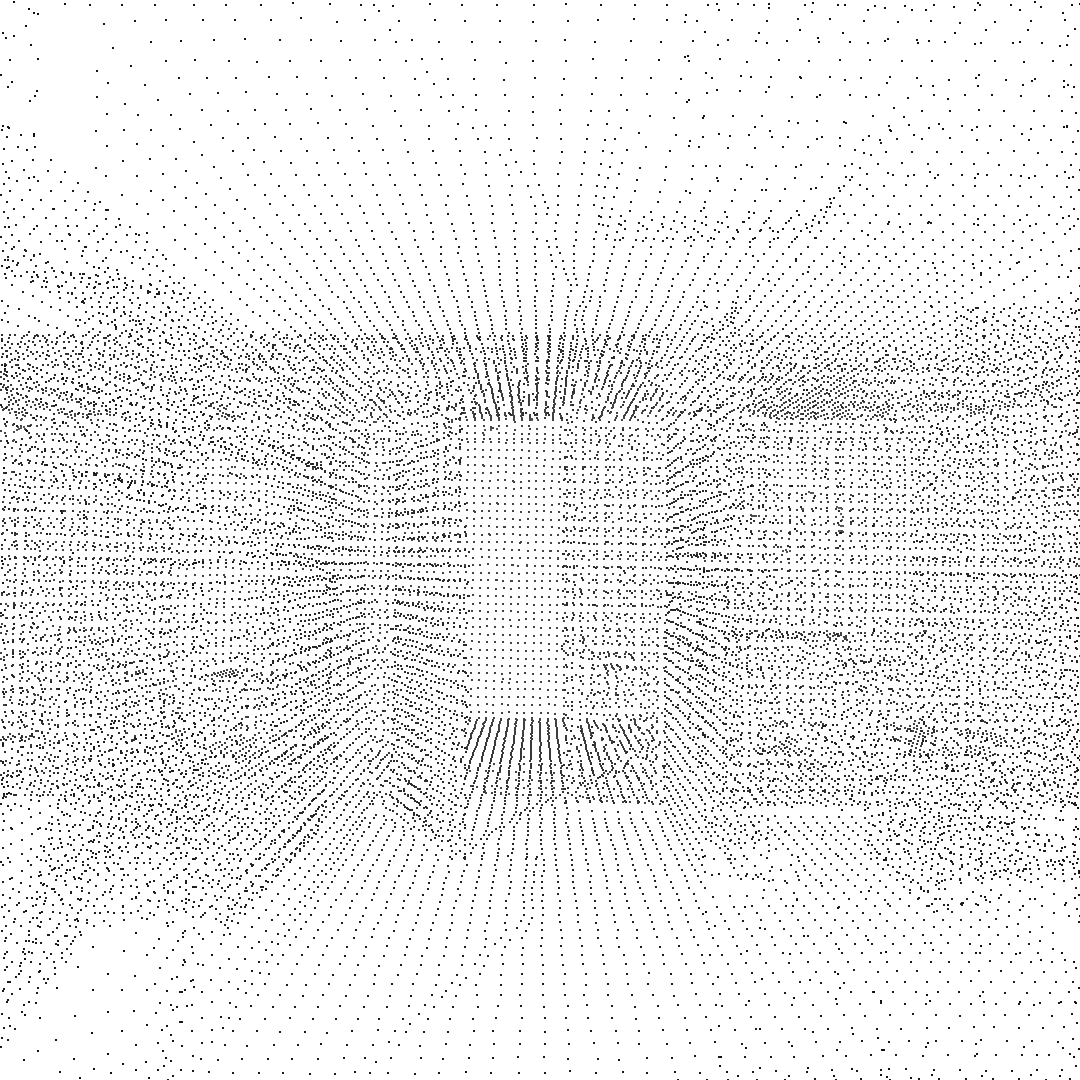}
    \caption{After removal}
    \label{fig:2d3ds_clean_after}
  \end{subfigure}
  \caption{Examples of IPR results. (a) Input images with one toy example and one 2D-3D-S example \cite{2D-3D-S}. (b) Corresponding point cloud submaps. (c) Submaps after removing invisible points from (b).}
  \label{fig:2d3ds}
\end{figure}

\textbf{Removal of invisible 3D points}.
Before extracting point cloud features, we exclude the parts of point cloud $P$ that are invisible in the query image $I$.
In order to identify occluded points, one intuitive idea is to keep only the closest 3D points that can be projected per image pixel to the depth map $D \in \mathbb{R}^{H \times W \times 1}$, which is constructed by projecting the 3D point cloud $P$ to the image frame.
However, due to much empty space in a 3D point cloud, points from invisible objects can be projected onto other pixels and blended together, making it difficult to identify occluded points.
To effectively deal with this problem, we use min-pooling and max-pooling together.
Our \textit{Invisible 3D Point Removal (IPR)} algorithm removes the points where the depth value does not change when applying min-pooling and max-pooling sequentially to the depth map as follows:
\begin{equation}
\text{IPR}(P)=P[\text{maxpool}(\text{minpool}(D)) = D].
\label{eq:mask}
\end{equation}
We set the kernel size of $s=9$ for both poolings. 
For each 3D point in the point cloud $P$, called a query point, the first min-pooling searches for the closer 3D points. 
Then the max-pooling checks whether those closer 3D points surround the query point.
This prevents 3D points from invisible objects from being projected onto other pixels and blended together due to the large amount of empty space in the 3D point cloud.
The time complexity of the IPR algorithm is $O(M + HWs^2)$, where $O(M)$ is for projecting the point cloud, $O(HWs^2)$ for building a depth map and pooling, and $O(M)$ for removing points.
\Cref{fig:2d3ds} shows some examples of invisible point removal.
The efficiency and effectiveness of our IPR algorithm can be seen in \Cref{tab:ablation_study} and discussed in \Cref{sec:ablation}. 

\textbf{3D point cloud descriptors}.
We voxelize the point cloud submaps using Minkowski Engine \cite{ME} to form the input tokens. 
After applying the IPR algorithm, we extract 3D local features using the Fast Point Transformer. 
We add two branches at the end of this network; one is followed by a NetVLAD layer that aggregates 3D point descriptors into a global descriptor, and the other is for 2D patch classification, which will be described in \Cref{sec:2d_patch_classification}.
Finally, we obtain $128$-dim local descriptors per 3D point and a $256$-dim global descriptor of the point cloud submap.

\subsection{2D patch classification} \label{sec:2d_patch_classification}

2D patch classification determines whether each 3D point is included in the image or not, and if yes, to which 2D patch in the image it belongs.
It can significantly reduce the search space to the patch-level during inference.
The number of classes to classify is the number of 2D patches plus one class indicating the 3D point does not belong to the image.
After concatenating the 3D point descriptors and the 2D patch descriptors, we apply an MLP with softmax to derive classification scores, as shown in \Cref{fig:model_details}.

\subsection{2D pixel features with positional encoding} \label{sec:positional_encoding}

In order to increase the number of inliers during matching the features between 2D pixels and 3D points, it is desired to have dense features in at least one modality.
Intuitively, the 2D image is better than the 3D point cloud for dense feature extraction, since the image is sampled based on a dense grid of pixels.
Assuming that the 2D patch-level features in the previous section are expressive enough to represent 2D pixel-level features,
we design a \textit{2D patch-to-2D pixel function} that returns the 2D pixel features from the corresponding 2D patch feature using positional encoding.
We use a simple two-layer MLP, pixel-MLP in \Cref{fig:model_details}, whose inputs are the 2D patch features $F_{\textrm{patch}}$ and the positional encoding $PE(r(\mathbf{x}))$ in 2D pixel coordinates $\mathbf{x}$.
$r$ is a function that converts $\mathbf{x}$ to coordinates within the patch containing $\mathbf{x}$, normalized to $[-1, 1]$.
We adopt the multi-dimensional positional encoding used in NeRF \cite{nerf} with the frequency range of $L=6$: 
\begin{equation}
  F_{\mathbf{x}} = MLP(F_{\textrm{patch}}, PE(r(\mathbf{x}))),
\end{equation}
\vspace{-22pt}
\begin{equation}
  PE(\mathbf{p}) = \{\textrm{sin} (2^f \pi \mathbf{p}), \textrm{cos} (2^f \pi \mathbf{p}) \ | \ f \in [0, L-1]\},
\end{equation}
where $\mathbf{p}=r(\mathbf{x})$.
Since there are multiple frequencies in the positional encoding, 2D pixel features are sufficiently distinctive from 2D patch features. 
To increase the generalizability, we use the coordinates obtained by projecting 3D points onto the image directly. 
It can learn features regardless of the pixel grid, which makes features expressed smoothly with continuous positional encoding.
However, for efficiency, we only extract features for each pixel coordinate and compare them with 3D point features at test. 

Our approach enables hierarchical matching; we first compare 3D point features with 2D patch features, and then with 2D pixel features within the 2D patch, to efficiently perform dense local matching. As a result, we can reduce the search space from $O(HW)$ to $O(\frac{HW}{g^2}+g^2)$ for a patch size $g$, and eventually to $O(\sqrt{HW})$ if $g=(HW)^{\frac{1}{4}}$.

\subsection{A differentiable PnP layer} \label{sec:diff_pnp}

Finally, a differentiable PnP solver learns weights $w_i$ to find the best 2D-3D correspondences for high-quality $6$-DoF pose estimation.
Specifically, the PnP solver is to find the optimal camera rotation matrix $\mathbf{R}^{*}$ and translation $\mathbf{t}^{*}$ that minimizes the sum of squared weighted reprojection errors of the 2D-3D correspondences ($\mathbf{x_i}$, $\mathbf{X_i}$).
\begin{equation}
\begin{aligned}
  & y^{*} = (\mathbf{R}^{*}, \mathbf{t}^{*})= \\
  & \operatorname*{argmin}_{y=(\mathbf{R}, \mathbf{t}) \in SE(3)} \frac{1}{2} \sum_i {\lVert \underbrace{w_i \circ\left(\pi(\mathbf{R} \mathbf{X_i} + \mathbf{t}) - \mathbf{x_i}\right)}_{f_i(y) \in \mathbb{R}^2} \rVert}^2 , 
\end{aligned}
\label{eqn:pnp}
\end{equation}
where $\circ$ is the element-wise product, $\pi$ is a function that projects 3D points of the camera coordinate system to 2D points of the image coordinate system based on the camera intrinsics $\mathbf{K}$, and $f_i(y)$ is the weighted reprojection error.
To this end, we adopt EPro-PnP \cite{epropnp} for end-to-end training. 

\subsection{Training} \label{sec:training}

Our loss function is defined by four terms of global matching loss, patch and pixel losses, and a PnP loss:
\begin{equation}
  L_{Total}= \alpha\mathcal{L}_{\text{Global}} + \beta\mathcal{L}_{\text{Patch}} + \gamma\mathcal{L}_{\text{Pixel}} + \mathcal{L}_{\text{KL}}.
\end{equation}
We set $\alpha=\beta=\gamma=0.5$ for experiments.

For global feature matching, 
we adopt a \textit{Triplet margin loss} \cite{triplet} defined as 
\begin{equation}
\mathcal{L}_{\text{Triplet}}\left(a, p, n\right)=\left[m+D\left(a, p\right)-D\left(a, n\right)\right]_{+},
\end{equation}
where $D\left(\cdot, \cdot\right)$ is the Euclidean distance, $[x]_{+}=\max (x, 0)$ is the hinge loss, and $m$ is a hyper-parameter for the margin (we set $m=0.4$).
$a$, $p$, and $n$ indicate anchor, positive, and negative, respectively.
We construct a set of training triplet tuples $\{I, P, N\}$, where $I$ is an image, and $P$ and $N$ respectively represent its positive point cloud submap and a set of negative submaps, for which we randomly sample submaps from the regions outside of $P$.
Then, our global feature matching loss is defined as
\begin{equation}
  \mathcal{L}_{\text{Global}}=\mathcal{L}_{\text{Triplet}}\left(G_I, G_P, G_N\right),
\end{equation}
where $G_I$, $G_P$, and $G_N$ are global features extracted from $I$, $P$, and $N$, respectively.

For the patch loss $\mathcal{L}_{\text{Patch}}$, we use the cross-entropy loss since the task is a 2D patch classification.

2D pixel feature learning also uses triplet margin loss.
For a 3D point feature $F_{point}$, $F_{\mathbf{x}}$ and $F_{\mathbf{x_n}}$ are respectively positive and negative features, and the loss is
\begin{equation}
  \mathcal{L}_{\text{Pixel}}=\mathcal{L}_{\text{Triplet}}\left(F_{point}, F_{\mathbf{x}}, F_{\mathbf{x_n}}\right).
\end{equation}
A negative pixel $\mathbf{x_n}$ is randomly sampled from the pixels that are at least $\frac{g}{2}$, half the patch size away from $\mathbf{x}$.


For training of the differentiable PnP solver, we view the PnP output as a pose distribution.
Letting $y$ as the estimated pose and $y_{gt}$ as the ground-truth pose:
the following KL divergence is minimized as a training loss.
\begin{equation}
\mathcal{L}_{\text{KL}}=\frac{1}{2} \sum_{i=1}\left\|f_i\left(y_{\mathrm{gt}}\right)\right\|^2+\log \int \exp \left( -\frac{1}{2} \sum_{i=1}\left\|f_i(y)\right\|^2 \right)  \hspace{-2pt}\mathrm{~d} y.
\end{equation}

\subsection{Inference} \label{sec:inference}

At inference, we predict the $6$-DoF camera coordinates of a query image using the point cloud submaps as a database.
We first extract 2D patch-level local features and a global descriptor as presented in \ref{sec:feature_extraction}. 
We retrieve the top $4$ 3D point cloud submaps from the database using the global descriptor.
The database is constructed by combining all of the training data from test regions that are not used during training, 
and the global descriptor and the 3D point descriptors are extracted from the point cloud submap with invisible points removed for the associated training image. 
Then, we perform 2D patch classification of each 3D point in the retrieved point cloud submaps to determine whether the 3D points of the candidates belong to which grid of the image, 
and calculate the 2D pixel coordinates of 3D points that are determined to belong to the image using positional encoding.
We put all 2D-3D correspondences into the differentiable PnP layer in \Cref{sec:diff_pnp}, 
and finally predict the $6$-DoF camera pose of the query image.
\Cref{fig:pipeline} illustrates the overall pipeline of our approach.

\section{Experiments}

\subsection{Benchmark datasets} \label{sec:datasets}

Existing benchmarks for image-based visual localization \cite{nuscenes1, nuscenes2, kitti, oxford} are often collected at a few different times at the same location.
This may lead to positional errors in the pose sensors and poor pose alignment between sequences, resulting in inaccurate 2D-3D correspondences. 
Therefore, we curate new benchmark datasets for the 2D image to 3D point cloud localization task. 

\begin{figure}[t]
  \centering
  \adjustbox{minipage=-0em,valign=t}{\subcaption{}\label{sfig:2d3ds_pc}}%
  \begin{subfigure}[t]{\dimexpr.5\linewidth-0.2em\relax}
  \centering
  \includegraphics[width=.75\linewidth,valign=t]{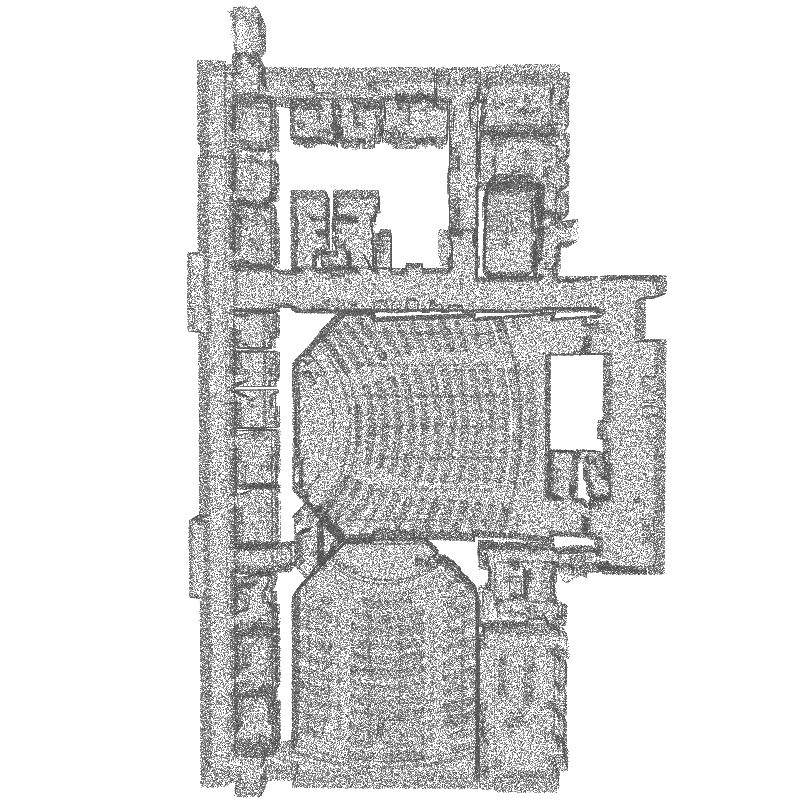}
  \end{subfigure}
  \centering
  \adjustbox{minipage=-0em,valign=t}{\subcaption{}\label{sfig:kitti_pc}}%
  \begin{subfigure}[t]{\dimexpr.5\linewidth-0.2em\relax}
  \centering
  \includegraphics[width=.75\linewidth,valign=t]{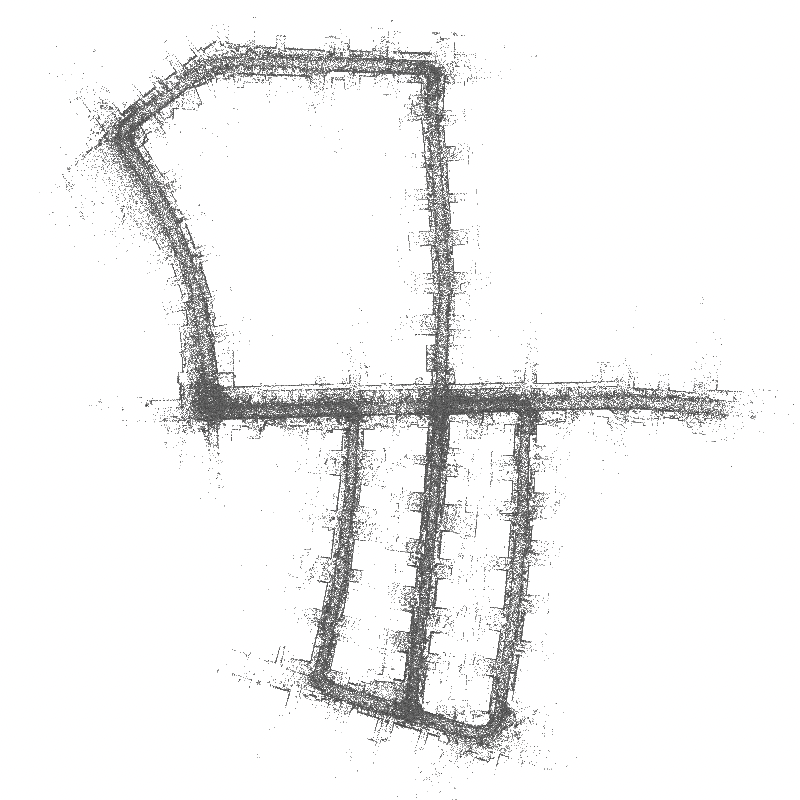}
  \end{subfigure}
  \caption{Global point cloud maps of (a) 2D-3D-S \cite{2D-3D-S} (Area 2) and (b) KITTI \cite{kitti} (Sequence 05).} 
  \label{fig:global_point_cloud}
\end{figure}

Our benchmark datasets are based on the indoor Stanford 2D-3D-Semantic (2D-3D-S) \cite{2D-3D-S} and the outdoor KITTI \cite{kitti} dataset, as shown in \Cref{fig:global_point_cloud}. 
The 2D-3D-S dataset contains $6$ large-scale indoor areas scanned by the MatterPort camera, 
and each area includes the global point cloud map provided in the Stanford large-scale 3D Indoor Spaces dataset (S3DIS) \cite{s3dis}.
We sample RGB images for training and testing so that the angles between the images taken at the same location do not differ by less than $30^\circ$ and $15^\circ$, respectively, following the InLoc \cite{inloc} setting.
For the coarse-to-fine matching approach, we provide the global point cloud maps along with small submaps.
We divide the global point cloud map into fixed-size submaps visible in the training image frames within $10$m and downsample to have $M=65,536$ points. 
We follow the $3$-fold cross-validation scheme of 2D-3D-S as shown in \Cref{tab:2d3ds_dataset_split},
which evaluates the performance of pose estimation for unseen places during training.
In each fold, the database for the test areas is constructed by collecting the training data for those areas that are not used during training. 
For example, the database for the Fold $3$ consists of the training data from Areas $2$, $4$, and $5$ combined.

The KITTI dataset captures large-scale outdoor environments by driving on the road in a medium-sized city of rural area.
There are $11$ sequences numbered $00$ to $10$ that have ground-truth poses, and we split the sequences $00$ to $08$ for training and the sequences $09$ to $10$ for testing.
Images are collected every $2$m from the trajectory with no preprocessing.
We construct a point cloud submap with the global coordinates by accumulating 3D LiDAR data and sampling the points visible in the image frame within $30m$,
and downsampling each submap to $M=65,536$ points. 
More details are described in the supplementary material.

\subsection{Metrics}

\begin{table}[t]
   \centering
   \resizebox{.9\linewidth}{!}{%
     \begin{tabular}{cc|cc|cc}
     \toprule
     \multirow{1}{*}{} & \multicolumn{1}{c|}{} & \multicolumn{2}{c}{Training} & \multicolumn{2}{|c}{Test} \\
     \multirow{1}{*}{Dataset} & \multicolumn{1}{c|}{Fold} & \multicolumn{1}{c}{Areas } & \multicolumn{1}{c}{\# of data}  & \multicolumn{1}{|c}{Areas} & \multicolumn{1}{c}{\# of data}\\
     \midrule
     \midrule
     \multirow{1}{*}{} & \multicolumn{1}{c|}{1} & \multicolumn{1}{c}{1, 2, 3, 4, 6} & \multicolumn{1}{c}{8,134} & \multicolumn{1}{|c}{5}  & \multicolumn{1}{c}{6,614} \\
     \multirow{1}{*}{2D-3D-S \cite{2D-3D-S}} & \multicolumn{1}{c|}{2} & \multicolumn{1}{c}{1, 3, 5, 6} & \multicolumn{1}{c}{6,482} & \multicolumn{1}{|c}{2, 4}  & \multicolumn{1}{c}{10,123} \\
     \multirow{1}{*}{} & \multicolumn{1}{c|}{3} & \multicolumn{1}{c}{2, 4, 5} & \multicolumn{1}{c}{7,218} & \multicolumn{1}{|c}{1, 3, 6}  & \multicolumn{1}{c}{8,511} \\
     
     \midrule
     \multirow{1}{*}{KITTI \cite{kitti}} & \multicolumn{1}{c|}{-} & \multicolumn{1}{c}{00 - 08} & \multicolumn{1}{c}{7,950} & \multicolumn{1}{|c}{09, 10}  & \multicolumn{1}{c}{1,052} \\
     \bottomrule
     \end{tabular}
     }
 \caption{The dataset configuration of 2D-3D-S \cite{2D-3D-S} and KITTI \cite{kitti}.}
 \label{tab:2d3ds_dataset_split}
 \end{table}

We evaluate two types of performances: the pose estimation of a query image (i) within the entire point cloud (\ie visual localization) and (ii) within a single point cloud (\ie image-to-point cloud registration).
For the visual localization task, we report the percentage of query images that are correctly localized within three thresholds of translation and rotation: ($0.1$m, $1.0^\circ$),  ($0.25$m, $2.0^\circ$), and ($1.0$m, $5.0^\circ$). 
This performance measurement follows the hardest evaluation method for the indoor environment of the \textit{Long-Term Visual Localization}\footnote{https://www.visuallocalization.net/benchmark/} challenge.
For the image-to-point cloud registration task, we report average Relative Translational Error (RTE (m)) and average Relative Rotation Error (RRE ($^\circ$)).

\subsection{Implementation details} \label{sec:implementation_details}

During training, we are given an image, the positive point cloud submap, a set of negative point cloud submaps, the camera pose, and the camera intrinsic matrix.
To shorten the training time, we organize a batch by sampling training images with the positive point cloud submap that can be used as negatives for each other. 
For 2D patch classification and 2D pixel feature learning, we generate ground-truth by projecting 3D points from the positive point cloud submap onto the image using the camera intrinsic matrix and the relative pose between the image and the point cloud.
If the 3D point is projected within the image frame, we specify the 2D patch number as a class number starting at $1$ and set it to the ground-truth along with the 2D pixel coordinates. 
Otherwise, we set the 2D patch number of the 3D point to the $0$-th class, indicating that it does not belong to the image frame. 
During training, these 2D patch numbers are used for 2D patch classification and the 2D pixel coordinates are used for triple tuple setup.
The loss function is minimized using Adam optimizer \cite{adam}, for which the initial learning rate is $10^{-4}$ and divided by $10$ at the epoch given in the LR scheduler steps.

\begin{table*}[t]
   \centering
   \resizebox{.95\textwidth}{!}{%
     \begin{tabular}{c|ccc|ccc|ccc|c}
     \toprule
     \multirow{1}{*}{} & \multicolumn{3}{c}{Fold 1} & \multicolumn{3}{|c}{Fold 2} & \multicolumn{3}{|c}{Fold 3} & \multicolumn{1}{|c}{Runtime} \\
     \multirow{1}{*}{Model} & \multicolumn{1}{c}{($0.1$, $1.0$)} & \multicolumn{1}{c}{($0.25$, $2.0$)} & \multicolumn{1}{c}{($1.0$, $5.0$)} & \multicolumn{1}{|c}{($0.1$, $1.0$)} & \multicolumn{1}{c}{($0.25$, $2.0$)} & \multicolumn{1}{c}{($1.0$, $5.0$)} & \multicolumn{1}{|c}{($0.1$, $1.0$)} & \multicolumn{1}{c}{($0.25$, $2.0$)} & \multicolumn{1}{c}{($1.0$, $5.0$)} & \multicolumn{1}{|c}{(sec)} \\
     
     \midrule
     \midrule
     \multirow{1}{*}{COLMAP \cite{colmap}} & \multicolumn{1}{c}{68.78} & \multicolumn{1}{c}{73.21} & \multicolumn{1}{c}{75.40} 
     & \multicolumn{1}{|c}{79.75} & \multicolumn{1}{c}{82.62} & \multicolumn{1}{c}{84.21} 
     & \multicolumn{1}{|c}{65.39} & \multicolumn{1}{c}{70.36} & \multicolumn{1}{c}{73.43} & \multicolumn{1}{|c}{9.377} \\
     
     \multirow{1}{*}{Kapture \cite{kapture}} & \multicolumn{1}{c}{72.80} & \multicolumn{1}{c}{77.49} & \multicolumn{1}{c}{79.56} 
     & \multicolumn{1}{|c}{81.39} & \multicolumn{1}{c}{84.88} & \multicolumn{1}{c}{86.58} 
     & \multicolumn{1}{|c}{66.96} & \multicolumn{1}{c}{72.28} & \multicolumn{1}{c}{74.53} & \multicolumn{1}{|c}{1.911} \\
     
     
     \multirow{1}{*}{SuperPoint \cite{superpoint} + NetVLAD \cite{netvlad}} & \multicolumn{1}{c}{81.49} & \multicolumn{1}{c}{83.26} & \multicolumn{1}{c}{84.23} 
     & \multicolumn{1}{|c}{88.57} & \multicolumn{1}{c}{89.89} & \multicolumn{1}{c}{90.37} 
     & \multicolumn{1}{|c}{78.86} & \multicolumn{1}{c}{81.20} & \multicolumn{1}{c}{82.25} & \multicolumn{1}{|c}{3.892} \\
     
     \multirow{1}{*}{HLoc \cite{hfnet, superglue}} & \multicolumn{1}{c}{84.14} & \multicolumn{1}{c}{86.50} & \multicolumn{1}{c}{87.86} 
     & \multicolumn{1}{|c}{90.47} & \multicolumn{1}{c}{92.10} & \multicolumn{1}{c}{92.86} 
     & \multicolumn{1}{|c}{80.84} & \multicolumn{1}{c}{83.80} & \multicolumn{1}{c}{85.18} & \multicolumn{1}{|c}{6.932} \\
     
     \multirow{1}{*}{D2-Net$^\dagger$ \cite{d2net} + NetVLAD} & \multicolumn{1}{c}{80.06} & \multicolumn{1}{c}{82.67} & \multicolumn{1}{c}{84.00} 
     & \multicolumn{1}{|c}{86.81} & \multicolumn{1}{c}{88.88} & \multicolumn{1}{c}{89.89} 
     & \multicolumn{1}{|c}{76.13} & \multicolumn{1}{c}{79.34} & \multicolumn{1}{c}{80.99} & \multicolumn{1}{|c}{5.815} \\
     
     \multirow{1}{*}{DSAC*\textsubscript{image-to-image} \cite{dsac_star}} & \multicolumn{1}{c}{0.00} & \multicolumn{1}{c}{0.00} & \multicolumn{1}{c}{0.29} 
     & \multicolumn{1}{|c}{0.16} & \multicolumn{1}{c}{1.56} & \multicolumn{1}{c}{11.41} 
     & \multicolumn{1}{|c}{0.74} & \multicolumn{1}{c}{5.18} & \multicolumn{1}{c}{19.84} & \multicolumn{1}{|c}{\textbf{0.738}} \\
     
     \midrule
     \multirow{1}{*}{PCLoc$^\dagger$ \cite{PCLoc}} & \multicolumn{1}{c}{76.34} & \multicolumn{1}{c}{80.99} & \multicolumn{1}{c}{83.87} 
     & \multicolumn{1}{|c}{76.97} & \multicolumn{1}{c}{80.19} & \multicolumn{1}{c}{83.92} 
     & \multicolumn{1}{|c}{74.26} & \multicolumn{1}{c}{77.97} & \multicolumn{1}{c}{80.41} & \multicolumn{1}{|c}{14.27} \\
     
     
     \multirow{1}{*}{DSAC*\textsubscript{RGB-D-to-RGB-D}$^\dagger$} & \multicolumn{1}{c}{0.00} & \multicolumn{1}{c}{0.00} & \multicolumn{1}{c}{0.26} 
     & \multicolumn{1}{|c}{0.00} & \multicolumn{1}{c}{0.52} & \multicolumn{1}{c}{8.94} 
     & \multicolumn{1}{|c}{0.86} & \multicolumn{1}{c}{3.75} & \multicolumn{1}{c}{13.54} & \multicolumn{1}{|c}{5.874} \\
     
     \midrule
     \multirow{1}{*}{\textbf{EP2P-Loc (Ours)}} & \multicolumn{1}{c}{\textbf{87.59}} & \multicolumn{1}{c}{\textbf{91.65}} & \multicolumn{1}{c}{\textbf{92.89}} 
     & \multicolumn{1}{|c}{\textbf{92.47}} & \multicolumn{1}{c}{\textbf{93.88}} & \multicolumn{1}{c}{\textbf{94.38}} 
     & \multicolumn{1}{|c}{\textbf{85.24}} & \multicolumn{1}{c}{\textbf{87.38}} & \multicolumn{1}{c}{\textbf{89.43}} & \multicolumn{1}{|c}{3.482} \\
     \bottomrule
     \end{tabular}
     }
 \caption{Experimental results of visual localization accuracy and runtime on the 2D-3D-S dataset \cite{2D-3D-S}. Please refer to \Cref{tab:2d3ds_dataset_split} for the folds. The threshold units are ($m$, $ ^\circ$). $^\dagger$ indicates models that use depth maps during training.}
 \label{tab:2d3ds_dataset_results}
 \end{table*}

 \begin{table}
   \centering
   \resizebox{.9\linewidth}{!}{%
     \begin{tabular}{c|c|c}
     \toprule
     \multirow{1}{*}{Model} & \multicolumn{1}{c}{RTE (m)} & \multicolumn{1}{|c}{RRE ($^\circ$)} \\
     
     \midrule
     \midrule
     \multirow{1}{*}{Direct Regression} & \multicolumn{1}{c}{4.94 $\pm$ 2.87} & \multicolumn{1}{|c}{21.98 $\pm$ 31.97} \\
     \multirow{1}{*}{Monodepth2 \cite{monodepth2} + USIP \cite{usip}} & \multicolumn{1}{c}{30.4 $\pm$ 42.9} & \multicolumn{1}{|c}{140.6 $\pm$ 157.8} \\
     \multirow{1}{*}{Monodepth2 + GT-ICP \cite{ICP1, ICP2}} & \multicolumn{1}{c}{2.9 $\pm$ 2.5} & \multicolumn{1}{|c}{12.4 $\pm$ 10.3} \\
     \multirow{1}{*}{2D3D-MatchNet \cite{2d3dmatchnet}$^\dagger$} & \multicolumn{1}{c}{752.5 $\pm$ 6053.3} & \multicolumn{1}{|c}{117.9 $\pm$ 52.1} \\
     \multirow{1}{*}{DeepI2P \cite{deepi2p}} & \multicolumn{1}{c}{3.28 $\pm$ 3.09} & \multicolumn{1}{|c}{7.56 $\pm$ 7.63} \\
     
     \midrule
     \multirow{1}{*}{\textbf{EP2P-Loc (Ours)}} & \multicolumn{1}{c}{\textbf{1.32 $\pm$ 1.13}} & \multicolumn{1}{|c}{\textbf{4.11 $\pm$ 5.46}} \\
     \bottomrule
     \end{tabular}
     }
 \caption{Experimental results of image-to-point cloud registration on the KITTI dataset \cite{kitti}. $^\dagger$ indicates the results of our implementation.}
 \label{tab:kitti_dataset_result}
 \end{table}

\subsection{Results of visual localization}

We compare our approach with six image-to-image localization methods, one image-to-RGB-D method \cite{PCLoc} and one RGB-D-to-RGB-D method \cite{dsac_star}.
The image-to-image baselines consist of a DSAC-based DSAC* \cite{dsac_star}, SfM methods such as COLMAP \cite{colmap} and Kapture \cite{kapture}, and the state-of-the-art local feature extraction methods including SuperPoint \cite{superpoint} + NetVLAD \cite{netvlad}, D2-Net \cite{d2net} + NetVLAD, and HLoc \cite{hfnet, superglue}.
The SfM baselines construct a 3D reference map using COLMAP on the database images.
The pipeline of local feature extraction methods consists of two steps: (i) retrieving the top $20$ nearest database images through NetVLAD and (ii) estimating pose by applying the local feature matching results to the PnP solver \cite{pnpsolver, epnp}.

\Cref{tab:2d3ds_dataset_results} summaries the results.
Our method outperforms existing image-to-image, image-to-RGB-D, or RGB-D-to-RGB-D localization approaches in all three-fold configurations with all three threshold settings.
Since the coordinate regression errors of the DSAC* are proportional to the area size, it becomes unreliable in large-scale localization. 
In addition, the RGB-D-to-RGB-D version of DSAC* also utilizes depth information of input images, but scores lower than the image-to-image version, since it highly downsamples the given depth map for efficiency but loses fine structure. 
The estimated pose of PCLoc \cite{PCLoc} depends on the top-$1$ candidate pose selected through the pose correction and pose verification, and does not fully utilize the given image, resulting in a low inlier problem. 
Another observation is that the image-to-image localization methods show better accuracy than DSAC* and PCLoc.
They use the SfM model to construct 2D-3D correspondences and are guided by SIFT \cite{sift} matching to triangulation.

\begin{table*}[t]
   \centering
   \resizebox{\textwidth}{!}{%
     \begin{tabular}{c|ccc|ccc|ccc|c|c|c}
     \toprule
     \multirow{1}{*}{} & \multicolumn{3}{c}{Fold 1} & \multicolumn{3}{|c}{Fold 2} & \multicolumn{3}{|c}{Fold 3} & \multicolumn{1}{|c}{} & \multicolumn{1}{|c}{} & \multicolumn{1}{|c}{Runtime} \\
     \multirow{1}{*}{Model} & \multicolumn{1}{c}{($0.1$, $1.0$)} & \multicolumn{1}{c}{($0.25$, $2.0$)} & \multicolumn{1}{c}{($1.0$, $5.0$)} & \multicolumn{1}{|c}{($0.1$, $1.0$)} 
     & \multicolumn{1}{c}{($0.25$, $2.0$)} & \multicolumn{1}{c}{($1.0$, $5.0$)} & \multicolumn{1}{|c}{($0.1$, $1.0$)} 
     & \multicolumn{1}{c}{($0.25$, $2.0$)} & \multicolumn{1}{c}{($1.0$, $5.0$)} & \multicolumn{1}{|c}{\# of points} & \multicolumn{1}{|c}{\# of inliers} & \multicolumn{1}{|c}{(sec)} \\
     \midrule
     \midrule
     
     \multirow{1}{*}{EP2P-Loc w/o IPR} & \multicolumn{1}{c}{\textbf{88.12}} & \multicolumn{1}{c}{90.33} & \multicolumn{1}{c}{92.75} 
     & \multicolumn{1}{|c}{91.57} & \multicolumn{1}{c}{\textbf{94.12}} & \multicolumn{1}{c}{\textbf{94.89}} 
     & \multicolumn{1}{|c}{83.23} & \multicolumn{1}{c}{86.36} & \multicolumn{1}{c}{88.29} & \multicolumn{1}{|c}{65.5K} & \multicolumn{1}{|c}{3.12K} & \multicolumn{1}{|c}{6.981} \\
     
     \multirow{1}{*}{EP2P-Loc w/ EPnP \cite{epnp}} & \multicolumn{1}{c}{85.34} & \multicolumn{1}{c}{87.51} & \multicolumn{1}{c}{90.23} 
     & \multicolumn{1}{|c}{90.23} & \multicolumn{1}{c}{92.34} & \multicolumn{1}{c}{93.78} 
     & \multicolumn{1}{|c}{83.56} & \multicolumn{1}{c}{84.96} & \multicolumn{1}{c}{87.13} & \multicolumn{1}{|c}{20.4K} & \multicolumn{1}{|c}{0.32K} & \multicolumn{1}{|c}{12.53} \\
     
     \midrule
     \multirow{1}{*}{\textbf{EP2P-Loc (Ours)}} & \multicolumn{1}{c}{87.59} & \multicolumn{1}{c}{\textbf{91.65}} & \multicolumn{1}{c}{\textbf{92.89}} 
     & \multicolumn{1}{|c}{\textbf{92.47}} & \multicolumn{1}{c}{93.88} & \multicolumn{1}{c}{94.38} 
     & \multicolumn{1}{|c}{\textbf{85.24}} & \multicolumn{1}{c}{\textbf{87.38}} & \multicolumn{1}{c}{\textbf{89.43}} & \multicolumn{1}{|c}{20.4K} & \multicolumn{1}{|c}{2.73K} & \multicolumn{1}{|c}{3.482} \\
     \bottomrule
     \end{tabular}
     }
 \caption{Ablation study on the 2D-3D-S dataset \cite{2D-3D-S}. The threshold units are ($m$, $ ^\circ$). The number of points is reported with the average value per point cloud submap in the training set. The number of inliers and runtime are reported with the average value per image in the test set, and runtime is measured with the overall pipeline including data loading to pose estimation.}
 \label{tab:ablation_study}
 \end{table*}

\subsection{Results of image-to-point cloud registration}

This task takes a query image and a single point cloud as inputs and estimates the relative pose between the image and the point cloud.
As baselines, we first select two state-of-the-art methods for 2D-3D registration: 2D3D-MatchNet \cite{2d3dmatchnet} and DeepI2P \cite{deepi2p}.
Since the code of 2D3D-MatchNet is not released, we implement it by ourselves. 
We also add three baselines of DeepI2P: 
(i) Direct Regression, (ii) Monodepth2 \cite{monodepth2} + USIP \cite{usip}, and (iii) Monodepth2 + GT-ICP.
Monodepth2 is the state-of-the-art depth estimation method, and USIP is the method to estimate the pose between the depth map and the point cloud.
GT-ICP indicates the Interactive Closest Point (ICP) method \cite{ICP1, ICP2} that uses the ground-truth (GT) relative pose for initialization.
These three baselines have unfair advantages over the other methods by using additional data or GTs.
Please refer to \cite{deepi2p} for more details of the baselines. 

As shown in \Cref{tab:kitti_dataset_result}, our method shows the state-of-the-art performance on the KITTI dataset \cite{kitti}.
2D3D-MatchNet shows the worst performance;
as pointed out in \cite{deepi2p}, 2D3D-MatchNet suffers from a low inlier rate in the correspondences, since it should learn to overcome the drastic discrepancy between image-based SIFT \cite{sift} and point cloud-based ISS \cite{iss} keypoints.
Moreover, the inlier ratio could further decrease in the KITTI dataset, which mostly consists of front-view images where similar patterns repeat and far buildings are shown relatively small. 
This requires extremely many iterations of PnP and RANSAC \cite{pnpsolver}, and thus its localization is unstable.
The superiority of our method over DeepI2P shows that our idea of explicitly calculating 3D points to 2D pixel coordinates can perform better than DeepI2P's grid classification of each 3D point.

\subsection{Ablation study} \label{sec:ablation}

 \begin{figure}
  \centering
   \includegraphics[width=0.65\linewidth]{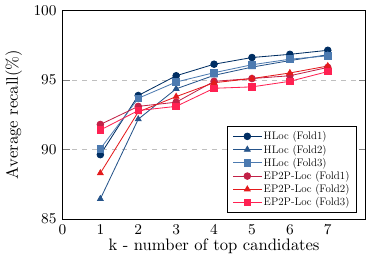}
   \caption{Comparison of average recalls for global matching on the 2D-3D-S dataset \cite{2D-3D-S}.}
   \label{fig:global_matching}
\end{figure}

\textbf{IPR algorithm.}
In the \Cref{tab:ablation_study}, we show the quantitative result of our IPR algorithm on the 2D-3D-S dataset \cite{2D-3D-S}.
Our IPR algorithm removes $68.9\%$ of the points within $10.6$ ms per submap during preprocessing. 
Even though the number of 3D points is reduced by a third, the number of inliers is reduced by only $12.5\%$, the performance is similar or better, and the runtime is reduced by $50.12\%$.

\textbf{Differentiable PnP solver.}
To show the effect of using a differentiable PnP solver in this task, we experiment with an EPnP solver \cite{epnp} with RANSAC \cite{pnpsolver}.
As shown in \Cref{tab:ablation_study}, the differentiable PnP solver outperforms an EPnP solver with the same 2D-3D correspondences, which weights the correspondences and efficiently estimating poses with a high number of inliers.

\subsection{Analysis} \label{sec:analysis} 

\textbf{Number of candidates to retrieval in global matching.}
We analyze the performance of the global descriptor with a different number of candidates.
We evaluate whether the RTE (m) for the top $k$ most similar candidates is lower than $0.1m$, 
and plot the recall curves of each fold of the 2D-3D-S dataset \cite{2D-3D-S}, as shown in \Cref{fig:global_matching}.
To show the effectiveness of our global descriptor, we compare our results with HLoc \cite{hfnet, superglue}, which is based on the coarse-to-fine approach as ours.
Our method performs comparably to image-to-image matching, and outperforms HLoc, especially for the top $1$. 
Moreover, our method can estimate a more accurate pose, as shown in \Cref{tab:2d3ds_dataset_results}.
While the recall value improves as $k$ increases, we construct the pipeline using $k=4$, considering the trade-off. 


\textbf{2D patch classification and 2D pixel coordinate calculation.}
\Cref{tab:2Dpatch2Dpixel} shows the accuracy of the 2D patch classification and the 2D pixel coordinate calculation on the 2D-3D-S and KITTI \cite{kitti} data.
The 2D-3D-S dataset, which is based on RGB-D data, has a more sophisticated data distribution than the KITTI dataset, which can lead to more accurate camera pose estimation. 

 \begin{table}
   \centering
   \resizebox{.85\linewidth}{!}{%
     \begin{tabular}{cc|c|c}
     \toprule
     \multirow{1}{*}{Dataset} & \multicolumn{1}{c}{Fold} & \multicolumn{1}{|c}{2D patch classification} & \multicolumn{1}{|c}{2D pixel calculation} \\
     
     \midrule
     \midrule
     \multirow{1}{*}{} & \multicolumn{1}{c}{1} & \multicolumn{1}{|c}{78.2} & \multicolumn{1}{|c}{41.5} \\
     \multirow{1}{*}{2D-3D-S \cite{2D-3D-S}} & \multicolumn{1}{c}{2} & \multicolumn{1}{|c}{82.3} & \multicolumn{1}{|c}{43.7} \\
     \multirow{1}{*}{} & \multicolumn{1}{c}{3} & \multicolumn{1}{|c}{70.8} & \multicolumn{1}{|c}{38.5} \\
     \midrule
     \multirow{1}{*}{KITTI \cite{kitti}} & \multicolumn{1}{c}{-} & \multicolumn{1}{|c}{63.8} & \multicolumn{1}{|c}{34.9} \\
     \bottomrule
     \end{tabular}
     }
 \caption{Results of 2D patch classification and 2D pixel calculation on the 2D-3D-S \cite{2D-3D-S}  and the KITTI \cite{kitti} datasets.}
 \label{tab:2Dpatch2Dpixel}
 \end{table}

 \textbf{Runtime comparison.}
\Cref{tab:2d3ds_dataset_results} presents that our model achieves superior performance while showing fast runtime compared to other state-of-the-art methods. 
DSAC* \cite{dsac_star} is a regression-based method, and Kapture \cite{kapture} uses GeM pooling \cite{gem} instead of NetVLAD \cite{netvlad}, resulting in faster runtime but showing inferior performance.
HLoc attains lower performance than our model but takes approximately twice the runtime.

\section{Conclusion}

To estimate the camera pose using a PnP solver for visual localization, it is necessary to obtain the 2D pixels of an image and the 3D point correspondences of a 3D reference map as input.
In this work, we propose a novel approach to solve the large-scale visual localization task by mitigating representational differences between 2D and 3D before feature extraction and finding all 2D-3D correspondences to increase the number of inliers, reducing memory usage and computational cost.
And we adopt an end-to-end trainable PnP solver, for the first time in this task, to learn to select good 2D-3D pairs for pose estimation utilizing the ground-truth of $6$-DoF camera pose during training. 
We experiment on our benchmark datasets based on large-scale indoor and outdoor environments,
and show that our method achieves the state-of-the-art performance compared to other previous visual localization and image-to-point cloud methods.

\textbf{Acknowledgement}.
We sincerely thank Junhyug Noh, Dongyeon Woo, Minui Hong, and Hyungsuk Lim for their constructive comments.
This work was supported by National Research Foundation of Korea (NRF) grant (No.2023R1A2C2005573) and Institute of Information \& Communications Technology Planning \& Evaluation (IITP) grant (No.2019-0-01082, No.2019-0-01309, No.2021-0-01343) funded by the Korea government (MSIT). Gunhee Kim is the corresponding author.

{\small
\bibliographystyle{ieee_fullname}
\bibliography{egbib}
}

\end{document}